\documentclass{article}

\usepackage[preprint]{neurips_2025}

\usepackage[utf8]{inputenc}
\usepackage[T1]{fontenc}
\usepackage{graphicx}
\usepackage{amsmath,amssymb,amsfonts}
\usepackage{booktabs,multirow}
\usepackage{adjustbox}
\usepackage{tabularx}
\newcolumntype{Y}{>{\raggedright\arraybackslash}X}
\usepackage{hyperref}
\usepackage{url}
\usepackage{xcolor}
\usepackage{placeins}
\usepackage{float}
\usepackage{microtype}
\usepackage{enumitem}
\setlist{nosep,leftmargin=*}
\title{ResPlan: A Large-Scale Vector-Graph Dataset of 17\,000 Residential Floor Plans}

\author{%
  Mohamed Abouagour \\
  Luddy School of Informatics, Computing, and Engineering \\
  Indiana University Bloomington \\
  \texttt{moabouag@iu.edu} \\
  \And
  Eleftherios Garyfallidis \\
  Luddy School of Informatics, Computing, and Engineering \\
  Indiana University Bloomington \\
  \texttt{elef@iu.edu} \\
}

\begin{document}

\maketitle

\begin{abstract}
ResPlan contains 17,000 residential floor plans with vector geometry, room-connectivity graphs, and metric-scale coordinates. Each plan annotates walls, doors, windows, and functional spaces (kitchens, bedrooms, bathrooms, balconies, and others) under a 17-class taxonomy, with polygons in pixel and meter coordinates. Four typed edges (\textit{via\_door}, \textit{adjacency}, \textit{direct}, \textit{via\_window}) accompany every plan, supporting graph-based generation and spatial reasoning. Compared with RPLAN~\citep{wu2019rplan} (raster-only, $\sim$6.7 rooms per plan, observed maximum of 8 functional rooms in our converted split) and MSD~\citep{vanengelenburg2024msdarxiv} (floor-plate-level, requiring extraction), ResPlan provides self-contained unit-level layouts averaging 8.1 functional rooms (9.2 graph nodes), spanning apartments, villas, and multi-wing residences. The release includes the dataset, loading/post-processing code, a canonical split, and baselines for three benchmark tasks. The dataset and code are publicly available.\footnote{\url{https://www.kaggle.com/datasets/resplan/resplan} and \url{https://github.com/m-agour/ResPlan}}
\end{abstract}

\begin{figure}[h!]
    \centering
    \includegraphics[width=\linewidth]{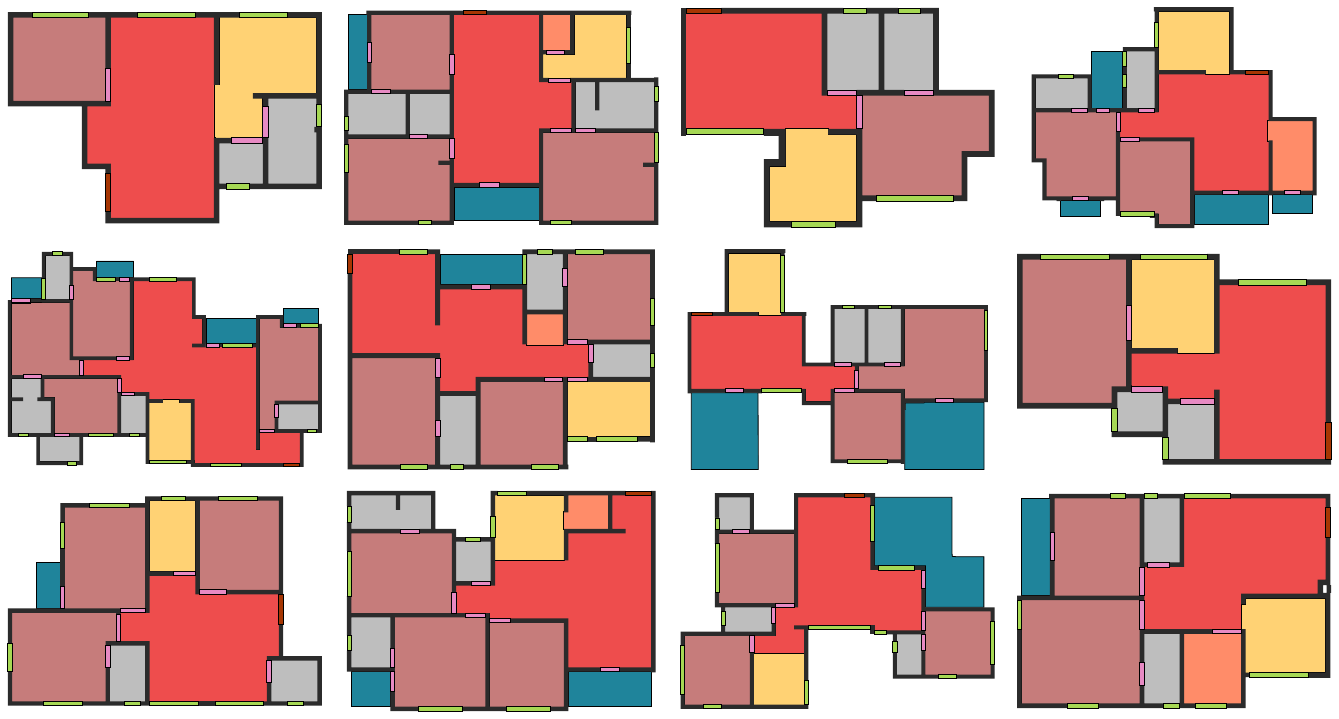}
    \vspace{-0.5em}
    \caption{Twelve floor plans from ResPlan, spanning small ($\le$6 rooms), medium (7--11), and large ($\ge$12) layouts, all rendered from the released vector geometry. Colors: \textcolor[HTML]{EE4D4D}{living area}, \textcolor[HTML]{C67C7B}{bedroom}, \textcolor[HTML]{BEBEBE}{bathroom}, \textcolor[HTML]{FFD274}{kitchen}, \textcolor[HTML]{FF8C69}{storage}, \textcolor[HTML]{1F849B}{balcony}; pink/green segments are doors/windows; dark strokes are walls.}
    \label{fig:main_resplan_example}
\end{figure}

%------------------------------------------------------------------------
\section{Introduction}
%------------------------------------------------------------------------
Floor plans are compact descriptions of how a home is organized: boundaries constrain space, room labels define function, and doors and windows determine how spaces connect. For learning systems that generate, label, or reason about layouts, these structural relationships matter as much as visual appearance. A dataset that only provides pixels can support image synthesis, but it cannot directly supervise metric geometry, room connectivity, or graph-conditioned reasoning.

Existing floor-plan datasets each provide part of this supervision. RPLAN~\citep{wu2019rplan} offers scale, with 80K plans, but only as low-resolution rasters and with relatively simple layouts ($\sim$6.7 rooms per plan; observed maximum of 8 functional rooms in the converted split used here). MSD~\citep{vanengelenburg2024msdarxiv} provides vector geometry and graphs for Swiss floor plates (18.9K units), but users must first extract individual apartments from building-level records. LIFULL~\citep{nii2017lifull} contains millions of raster images but only a small structured subset, while CubiCasa5K~\citep{kalervo2019cubicasa5k} provides 5K SVG plans with rich annotations but limited scale. The result is a practical trade-off: researchers can choose scale, vector geometry, or topology, but not all three in a self-contained, unit-level dataset. This trade-off is most visible for larger homes, where 34.6\% of ResPlan test plans contain more functional rooms than the converted RPLAN split's observed maximum (Section~\ref{sec:analysis}).

ResPlan is designed to remove this trade-off. It contains 17,000 residential floor plans, from apartments to multi-wing homes, with the visual diversity shown in Figure~\ref{fig:main_resplan_example}. Each plan includes polygonal geometry with consistent wall thickness, 17 semantic categories, metric-scale coordinates, and a room-connectivity graph with four typed edges (\texttt{via\_door}, \texttt{adjacency}, \texttt{direct}, \texttt{via\_window}). The release also provides code for loading, rendering, graph construction, validation, and post-processing, plus a canonical split (80/10/10) and baselines for three benchmark tasks, so the dataset can be used both as a resource and as a reference benchmark.

\begin{table}[t]
\centering
\caption{Comparison of ResPlan with existing floor plan datasets: RPLAN~\citep{wu2019rplan}, MSD~\citep{vanengelenburg2024msdarxiv}, LIFULL~\citep{nii2017lifull}, and CubiCasa5K~\citep{kalervo2019cubicasa5k}.}
\label{tab:dataset_comparison}
\begin{adjustbox}{max width=\linewidth}
\begin{tabular}{@{}llllll@{}}
\toprule
\textbf{Feature}           & \textbf{ResPlan} & \textbf{RPLAN} & \textbf{MSD} & \textbf{LIFULL} & \textbf{CubiCasa5K} \\
\midrule
\# Samples            & $\sim$17,000       & $\sim$80,000      & 5,372 (18.9k units) & $\sim$5M raster ($\sim$124k vector\textsuperscript{*}) & 5,000            \\
Source               & Online real estate listings & Chinese real estate & Swiss dwellings    & Japanese real estate & Multiple countries \\
Format               & Vector + Graph    & Raster (256$\times$256) & Vector + Graph  & Raster (+ R2V vectors)    & Raster + Vector (SVG) \\
Curation             & Cleaned, Filtered & Manual            & Cleaned, Filtered & Licensed (NII)      & Manual (+furniture)   \\
Vector Native        & Yes      & No (converted)   & Yes              & No                & Yes               \\
Unit-Level Plans     & Yes      & Yes             & No (extraction needed) & Yes             & Yes               \\
Graph Connectivity   & Yes      & No              & Yes            & No                & No                \\
Typed Edge Labels    & Yes (4 types) & No          & Yes (3 types)  & No                & No                \\
Avg Rooms/Plan       & 8.1      & $\sim$6.7        & $\sim$8.8 per unit & Varies            & --            \\
Max Rooms/Plan (obs.) & 22       & 8                & --               & --                & --                \\
Metric Scale         & Yes (meters)   & No                & Partial          & No                & No                \\
Canonical Splits     & Yes (80/10/10) & No         & Yes            & No                & No                \\
License              & CC BY 4.0         & Custom (research) & CC BY 4.0       & Restricted (NII)  & CC BY-NC-SA 4.0   \\
\bottomrule
\end{tabular}
\end{adjustbox}
\par\smallskip
{\footnotesize \textsuperscript{*}LIFULL HOME's contains $\sim$5M raster floor-plan images; only the $\sim$124k vectorized by Liu et al.~\citep{liu2017raster} are commonly used in structured pipelines.}
\end{table}

ResPlan provides a benchmark for structured floor-plan learning and analysis. Table~\ref{tab:dataset_comparison} summarizes the comparison with existing datasets; a quantitative comparison is given in Section~\ref{sec:analysis}.

\begin{figure}[t]
    \centering
    \includegraphics[width=0.75\linewidth]{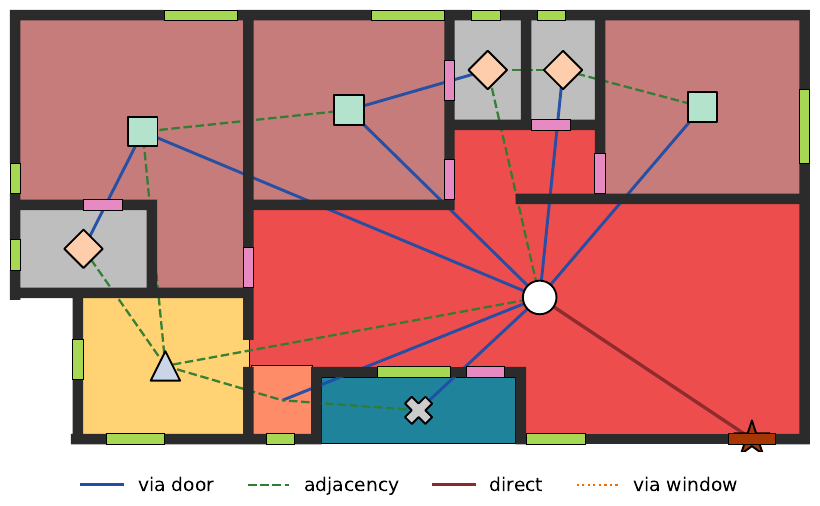}
    \caption{Graph-based representation of floor plans in the ResPlan dataset. Rooms are represented as nodes with semantic and geometric attributes, while edges encode adjacency relationships classified as \textit{via\_door}, \textit{adjacency}, \textit{direct}, or \textit{via\_window}.}
    \label{fig:resplan_graph_representation}
\end{figure}

%------------------------------------------------------------------------
\section{Related Work}
%------------------------------------------------------------------------

\paragraph{Raster Floor-Plan Datasets.}
RPLAN~\citep{wu2019rplan} is the most widely used floor-plan dataset, containing 80,788 annotated plans rendered as $256{\times}256$ raster images. Despite its scale, RPLAN predominantly consists of simple, single-unit layouts, and its raster format introduces irregular door geometries that complicate precise geometric analysis. LIFULL~\citep{nii2017lifull} offers millions of Japanese floor-plan images, but only a small subset has been converted to structured form. CubiCasa5K~\citep{kalervo2019cubicasa5k} supplies 5,000 SVG plans with rich annotations but at limited scale. These raster-first resources are valuable for recognition and segmentation, but they do not directly provide metric unit-level geometry paired with typed room-connectivity graphs.

\paragraph{Vector and Graph-Based Datasets.}
The Modified Swiss Dwellings (MSD) dataset~\citep{vanengelenburg2024msdarxiv} provides 5,372 floor plans (18,900 dwelling units) in vector, raster, and graph modalities with three edge types (\textit{passage}, \textit{door}, \textit{front-door}). ResPlan provides 17,000 self-contained residential units with metric geometry and four typed relations (window-mediated, door-mediated, front-door, and wall-adjacent), directly supporting generation, graph prediction, and search over functionally similar layouts.

\paragraph{Graph-Conditioned Generation.}
An increasing number of generative models condition layout synthesis on graph structures. House-GAN++~\citep{nauata2021houseganplusplus} and Graph2Plan~\citep{hu2020graph2plan} use room-connectivity graphs as input constraints, while diffusion methods such as HouseDiffusion~\citep{shabani2023houseDiffusion} and GSDiff~\citep{hu2025gsdiff} further demonstrate the utility of structured graph inputs. Figure~\ref{fig:resplan_graph_representation} illustrates ResPlan's representation: rooms are graph nodes, and edge labels distinguish access through doors/windows from direct contact and wall adjacency. Such methods need paired room geometry and connectivity graphs, which ResPlan provides natively with four typed edge labels.

%------------------------------------------------------------------------
\section{Dataset Description}
\label{sec:dataset}
%------------------------------------------------------------------------

\paragraph{Data Collection and Processing.}
ResPlan was built by converting rendered, textured listing images into annotated vectors. Source plans come from public real-estate listings rendered as stylized rasters: rooms use category-specific color/texture swatches (e.g., wood-tone bedrooms, tiled bathrooms/balconies), walls are thick dark strokes, and labels/dimensions are printed on the image (e.g., ``Bedroom~15'9''$\times$11'''). Conversion has three stages. First, color/texture segmentation separates rooms from structural strokes, while OCR over labels adds semantic cues. Second, connected components are traced into room, wall, door, and window polygons; classes combine visual cues and OCR, and listed gross area anchors metric scale. Third, post-processing repairs or discards invalid fragments, removes slivers/disconnected pieces, reconciles near-touching room boundaries, aligns openings to the nearest wall band, and normalizes wall thickness. The initial collection yielded approximately 27,000 raw plans; automated quality checks discarded layouts with degenerate geometries, missing structural elements, disconnected interiors, or ambiguous labels, leaving 17,000 plans (37\% rejection). Heterogeneous labels such as \emph{Master Bedroom}, \emph{Guest Room}, or \emph{Kids Room} are normalized into a 17-category schema.\footnote{The 17 categories comprise functional spaces (\texttt{living}, \texttt{bedroom}, \texttt{bathroom}, \texttt{kitchen}, \texttt{balcony}, \texttt{storage}, \texttt{stair}, \texttt{garden}, \texttt{parking}, \texttt{pool}), structural elements (\texttt{wall}, \texttt{door}, \texttt{window}, \texttt{front\_door}), and boundary layers (\texttt{inner}, \texttt{land}, \texttt{neighbor}).} Door-to-wall alignment is verified by automatic geometric checks plus author-led visual spot-checks. On a stratified 500-plan audit, 83.1\% of room polygons fall within plausible per-class area ranges (per-class 81.9--86.0\%); all sampled plans are graph-connected and no bathroom is graph-isolated. We release the derived dataset with loading, rendering, graph construction, validation, metric conversion, and post-processing code; source collection and OCR/vectorization scripts are not redistributed.

\paragraph{Data Provenance and Ethics.}
The source plans originate from online real-estate listing platforms serving South Asian residential markets. Platform identities are withheld to comply with platform access terms (LIFULL~\citep{nii2017lifull} convention); collection used only public, non-paywalled listing pages with no rate-limit, login-wall, or other access-control bypassing, after platform terms-of-service review. The released dataset consists entirely of derived geometric representations (Shapely~\citep{shapely2024} polygons and NetworkX graphs); no original drawings, images, listing text, prices, addresses, or personally identifiable information are retained. The transformation removes all copyrightable artistic expression, preserving only factual geometric topology. The data is distributed under the CC~BY~4.0 license; a full datasheet~\citep{gebru2021datasheets} is included in the supplementary material. A public takedown policy (\texttt{TAKEDOWN.md}) accompanies the release: rights holders may request removal of specific plan IDs, with acknowledgement within 7 days and removal in the next release within 30 days; removed IDs are tracked so downstream users can update their splits.

\paragraph{Release, Hosting, and Long-Term Availability.}
The dataset is hosted on Kaggle Datasets and mirrored with a permanent DOI. A Croissant~\citep{akhtar2024croissant} metadata file accompanies the release, so the dataset is discoverable and loadable by tools that consume Croissant. Code, trained baseline checkpoints, and a reproduction script (\texttt{reproduce.sh}) that regenerates every table and figure from the released dataset are provided in a public code repository with a pinned Docker image. The canonical split file (\texttt{split.json}) is versioned alongside the data; all reported numbers are tied to split-v2. The dataset is at \url{https://www.kaggle.com/datasets/resplan/resplan} and the code at \url{https://github.com/m-agour/ResPlan}.

\paragraph{Data Format and Schema.}
Each ResPlan sample is stored as a Python dictionary whose keys are semantic categories (e.g., \texttt{bedroom}, \texttt{bathroom}, \texttt{wall}, \texttt{door}) and whose values are Shapely polygon or multi-polygon objects in plan-view coordinates. If a category is absent in a given plan, its value is an empty geometry (\texttt{MULTIPOLYGON EMPTY}), so downstream pipelines never require special-case checks for missing keys.

Room polygons are grouped by function; structural elements (walls, doors, windows) share a common thickness (\texttt{wall\_depth}). The primary entrance (\texttt{front\_door}) is isolated as a distinct element. An \texttt{inner} polygon encloses inhabitable space, and a \texttt{neighbor} layer records party walls---shared boundary walls between the unit and adjacent properties---enabling exterior/interior wall disambiguation via boolean set operations.

\paragraph{Metadata and Splits.}
Every record includes a gross floor area in square meters and a numeric identifier. The canonical train/validation/test split is 13,736/1,632/1,632 plans (approximately 80/10/10), stratified by bedroom count to ensure proportional representation across subgroups. Split indices are distributed as a JSON file (\texttt{split.json}) mapping plan IDs to partitions.

\paragraph{Metric-Scale Coordinates.}
Every plan is also available in real-world meters, converted using the listed gross floor area to derive a meters-per-pixel scale factor applied via Shapely's \texttt{affine\_transform}. Each record stores the scale factor, wall thickness in meters, and per-room areas. Across the dataset, median plan width is 12.4\,m and median wall thickness is 21\,cm.

\paragraph{Graph Construction.}
Each plan is converted into a room-connectivity graph whose nodes correspond to six primary spaces (\texttt{bedroom}, \texttt{bathroom}, \texttt{kitchen}, \texttt{living}, \texttt{balcony}, \texttt{front\_door}), each storing a semantic label, polygon geometry, and area. Secondary spaces are present as geometry-only categories. Edges encode four types: \textit{via\_door} and \textit{via\_window} (two rooms bridged by a door or window), \textit{adjacency} (rooms sharing a wall, detected via buffered polygon overlap), and \textit{direct} (front-door nodes linked to rooms they physically touch). Polygon ordering is fixed, enabling $O(1)$ cross-referencing between graph nodes and geometries. Task~1 uses the five room-type nodes as labels; \texttt{front\_door} is retained as a graph node but excluded from the 5-class label set. Figure~\ref{fig:resplan_graph_representation} illustrates the graph representation.

%------------------------------------------------------------------------
\section{Dataset Analysis}
\label{sec:analysis}
%------------------------------------------------------------------------

\begin{figure}[t]
    \centering
    \includegraphics[width=\linewidth]{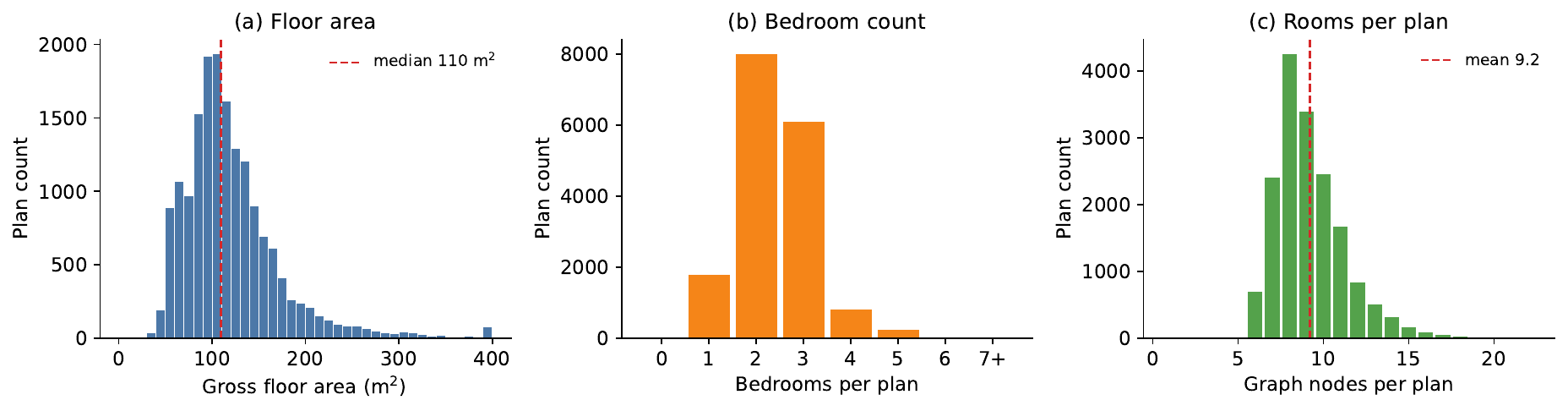}
    \caption{Summary statistics of the ResPlan dataset: (a)~floor area distribution (median 110~m$^2$); (b)~bedroom count distribution; (c)~graph nodes (rooms) per plan.}
    \label{fig:dataset_analysis}
\end{figure}

Figure~\ref{fig:dataset_analysis} presents the key distributional properties. Floor areas range from under 10~m$^2$ to over 700~m$^2$ (median 110~m$^2$, mean 121~m$^2$, $\sigma{=}55$~m$^2$). Two-bedroom plans are the most common (47\%), followed by three-bedroom (36\%); 6.4\% have four or more bedrooms. The room-connectivity graphs average 9.2 nodes and 12.9 edges per plan. Edge-type distribution: \textit{via\_door} 54.2\%, \textit{adjacency} 35.2\%, \textit{direct} 7.6\%, \textit{via\_window} 3.0\%. All but one graph is connected. A total of 125 plans (0.7\%) lack a kitchen polygon.

\noindent\textbf{Geometric statistics.} Across 533,788 polygons, median vertex count is 4 (53.8\% rectangles; 9.5\% L-shaped, 6 vertices), and plan aspect ratio averages 1.40 (median 1.33). Edge fractions are \textit{via\_door} $51.3{\pm}10.1$\%, \textit{adjacency} $32.5{\pm}10.8$\%, \textit{direct} $8.8{\pm}4.0$\%, and \textit{via\_window} $7.4{\pm}7.0$\%.

\begin{figure}[t]
    \centering
    \includegraphics[width=0.82\linewidth]{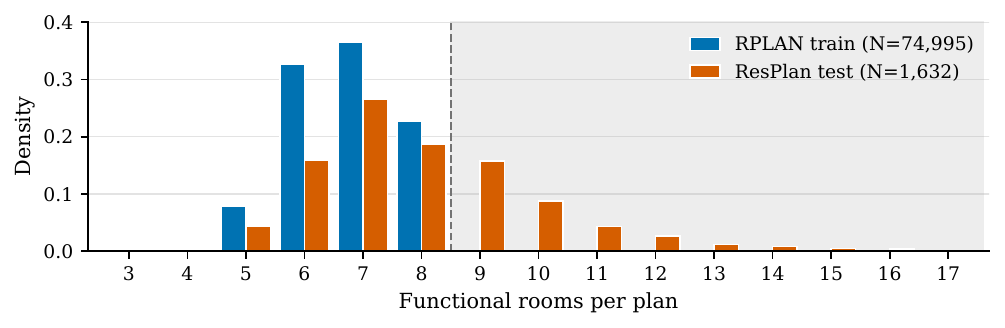}
    \caption{Functional-room-count distribution of the converted RPLAN training split (blue, N=74{,}995) vs.\ the ResPlan test split (orange, N=1{,}632), using the five shared room types. The observed RPLAN maximum is 8 rooms per plan; 34.6\% of ResPlan test plans exceed this limit.}
    \label{fig:coverage}
\end{figure}

\noindent\textbf{Scale and coverage.} At 17,000 plans, ResPlan is substantially larger than CubiCasa5K and the 5,372 floor-plate records in MSD, while providing self-contained unit-level geometry, typed graphs, and metric scale. Figure~\ref{fig:coverage} compares functional-room counts (bedroom, bathroom, kitchen, living, balcony) against the converted RPLAN training split used in our transfer experiment: that split has an observed maximum of eight rooms, while \textbf{34.6\% of ResPlan test plans lie outside this support}, extending to 17 rooms. Section~\ref{sec:task2} evaluates generation in this larger-layout regime.

%------------------------------------------------------------------------
\section{Benchmark Tasks}
\label{sec:tasks}
%------------------------------------------------------------------------
Three benchmark tasks cover semantic, geometric, and topological reasoning: semantic room labeling of graph nodes (accuracy, macro/weighted F1); constrained floor-plan generation from a boundary, room list, and graph (boundary IoU, room-count accuracy, adjacency satisfaction, FID); and plan-to-graph extraction with typed edges (edge precision/recall/F1 and type accuracy).

%------------------------------------------------------------------------
\section{Baseline Experiments}
\label{sec:baselines}
%------------------------------------------------------------------------
To demonstrate the utility of ResPlan and establish initial benchmarks, methods are evaluated on the three tasks defined above, together with a cross-dataset transfer experiment against RPLAN. The canonical split is used throughout; all metrics are computed on the 1{,}632-plan test partition.

\subsection{Task~1: Semantic Room Labeling}
Each room node in the connectivity graph must be classified into one of five categories: \textit{bedroom}, \textit{bathroom}, \textit{kitchen}, \textit{living}, or \textit{balcony}.

\paragraph{Methods.}
Five baselines are compared: a rule-based decision tree (area, edge-type degree, area ratio); a random forest (200 trees, depth 20, 11 node features); gradient boosting (200 trees, depth 6, same features); a 3-layer GCN~\citep{kipf2017gcn} (128 hidden, BN, dropout 0.3, Adam, 500 epochs); and a 3-layer GraphSAGE~\citep{hamilton2017graphsage} with the same setup but concatenation-based aggregation.

\begin{table}[t]
\centering
\caption{Baseline results for semantic room labeling (Task~1) on the ResPlan test set. GNN $\pm$ values are standard deviations over 3~random seeds; tree-based methods are deterministic.}
\label{tab:baselines}
\small
\begin{tabular}{@{}lccc@{}}
\toprule
\textbf{Method} & \textbf{Accuracy} & \textbf{Macro F1} & \textbf{Weighted F1} \\
\midrule
Rule-based (area + connectivity)  & 0.840 & 0.821 & 0.835 \\
GCN (3-layer, 128-dim)             & 0.739{\scriptsize$\pm$.004} & 0.765{\scriptsize$\pm$.005} & 0.741{\scriptsize$\pm$.005} \\
Gradient Boosting                 & 0.908 & 0.901 & 0.907 \\
Random Forest                     & 0.914 & 0.908 & 0.914 \\
GraphSAGE (3-layer, 128-dim)       & \textbf{0.951}{\scriptsize$\pm$.001} & \textbf{0.950}{\scriptsize$\pm$.001} & \textbf{0.951}{\scriptsize$\pm$.001} \\
\bottomrule
\end{tabular}
\end{table}

\paragraph{Results.}
Table~\ref{tab:baselines} summarizes the results. GraphSAGE achieves the highest accuracy (95.1\%) and macro F1 (0.950), outperforming tree-based methods (RF 91.4\%, GB 90.8\%) and GCN (73.9\%), with the difference statistically significant ($p < 10^{-21}$, Wilcoxon signed-rank). Per-class F1 ranges from 0.998 (living) to 0.873 (balcony), with the largest gains over tree methods on bathrooms (0.944 vs.\ 0.908) and kitchens (0.953 vs.\ 0.884), where graph context disambiguates rooms with overlapping sizes.

\subsubsection{Ablation Studies}
\label{sec:ablation}

\paragraph{Feature ablation.}
Table~\ref{tab:feat_ablation} shows GraphSAGE accuracy under progressively reduced feature sets. The full 11-feature set achieves 95.0\%. Removing typed-edge degrees drops accuracy by 2.3~pp. Area alone reaches 87.2\%, while a structure-only baseline (no node features) drops to 34.0\%, confirming that node attributes are necessary.

\begin{table}[t]
\centering
\caption{Feature ablation for GraphSAGE on Task~1. ``No edge-type degrees'' removes the three typed-edge degree features (via\_door, adjacency, via\_window), leaving 8 generic features. Mean$\pm$std over 3~seeds.}
\label{tab:feat_ablation}
\small
\begin{tabular}{@{}llcc@{}}
\toprule
\textbf{Feature set} & \textbf{Dim} & \textbf{Accuracy} & \textbf{Macro F1} \\
\midrule
All features                 & 11 & \textbf{0.950$\pm$0.001} & \textbf{0.949$\pm$0.001} \\
No edge-type degrees         &  8 & 0.927$\pm$0.000 & 0.922$\pm$0.000 \\
Area + ratio + centroid      &  4 & 0.916$\pm$0.001 & 0.909$\pm$0.001 \\
Area only                    &  1 & 0.872$\pm$0.001 & 0.862$\pm$0.000 \\
Structure only (no features) &  1 & 0.340$\pm$0.000 & 0.102$\pm$0.000 \\
\bottomrule
\end{tabular}
\end{table}

Among GNN architectures (3 layers, 128 hidden), GraphSAGE (95.1\%) outperforms GIN (83.9\%), GCN (73.8\%), and GAT (59.6\%).

\subsubsection{Cross-Dataset Transfer}
\label{sec:cross}
To quantify the domain gap between ResPlan and other vector floor-plan resources, GraphSAGE is trained on one dataset and evaluated on the others, using a reduced 8-feature set common to all three (room area, total degree, neighbor area statistics, area ratio, centroid position). RPLAN and MSD room categories are mapped to the same five classes (see supplementary); MSD apartments are extracted by grouping the v2 release's residential areas by \texttt{apartment\_id} with adjacency from polygon contact.

\begin{table}[t]
\centering
\caption{Cross-dataset transfer accuracy (GraphSAGE, 8 common features). Diagonal entries are in-domain; off-diagonal entries measure transfer. MSD rows use 18,002 residential apartments from the v2 release.}
\label{tab:cross}
\small
\begin{tabular}{@{}lcc@{}}
\toprule
\textbf{Train $\rightarrow$ Test} & \textbf{Accuracy} & \textbf{Macro F1} \\
\midrule
ResPlan $\rightarrow$ ResPlan & 0.926 & 0.920 \\
RPLAN $\rightarrow$ RPLAN       & 0.907 & 0.898 \\
MSD $\rightarrow$ MSD           & 0.818 & 0.810 \\
\midrule
RPLAN $\rightarrow$ ResPlan     & 0.582 & 0.578 \\
ResPlan $\rightarrow$ RPLAN     & 0.649 & 0.659 \\
MSD $\rightarrow$ ResPlan       & 0.190 & 0.152 \\
ResPlan $\rightarrow$ MSD       & 0.322 & 0.287 \\
\bottomrule
\end{tabular}
\end{table}

Table~\ref{tab:cross} shows substantial gaps to RPLAN (34.4/25.8~pp) and larger gaps to MSD (70.6/49.6~pp): despite both being residential vector datasets, ResPlan averages 8.1 functional rooms vs.\ RPLAN's 6.7, has lower graph density (0.35 vs.\ 0.53), more bathrooms (30.7\% vs.\ 17.9\%), and a smaller living-room fraction (35.9\% vs.\ 50.0\%); MSD's Swiss corridor-mediated layouts diverge further still, so ResPlan is complementary to both. RPLAN pre-training plus fine-tuning helps only at 1\% training data (+0.4~pp) and vanishes at $\geq$5\%, confirming the gap is structural.

\subsection{Task~2: Constrained Floor Plan Generation}
\label{sec:task2}
Given a target room-count vector and adjacency graph, the goal is to produce a floor plan that satisfies these constraints. Both retrieval-based and learned generative baselines are evaluated.

Baselines include random, count-matched, and count+graph retrieval, plus four learned methods. Two conditional VAEs predict room boxes $(c_x, c_y, w, h)$: CVAE uses room counts and boundary aspect ratio, while CVAE+Graph adds a 15-dimensional adjacency feature (both $\beta$-VAE, $\beta{=}0.1$, 128-dim latent, early stopping). A 3-layer layout GNN (13K parameters) predicts boxes by message passing, and an unconditional DDPM (UNet, dim multipliers $1,2,4,8$; $128{\times}128$; 100k steps, AMP, one L40S) assesses image-space quality independent of conditioning. Training rasters render six room categories plus walls; doors/windows are not separate channels, but the DDPM learns plausible openings as wall gaps.

GSDiff~\citep{hu2025gsdiff}, HouseDiffusion~\citep{shabani2023houseDiffusion}, and House-GAN++~\citep{nauata2021houseganplusplus} require nontrivial adaptation (wall-junction extractor and 17-category embedding for GSDiff; rectangular-box assumption for the others, but only 53.8\% of ResPlan polygons are rectangles); the released split and reproduction script support this follow-up.

\paragraph{FID protocol.}\label{sec:task2_protocol}
Bounding-box outputs from CVAE, CVAE+Graph, and Layout GNN are rasterized through a boundary-partition renderer: each inner-polygon pixel is assigned to the nearest predicted box (L$_\infty$ distance, ties by center distance and area), producing a non-overlapping partition. Applying the same renderer to ground-truth boxes yields the conditional \emph{partition-render} Fr\'echet Inception Distance (FID$_\text{part.}$). The unconditional DDPM outputs raster pixels directly and is evaluated against natively rasterized plans (FID$_\text{raster}$), so its FID is not directly comparable to Table~\ref{tab:gen_baselines}'s partition values.

\begin{table}[t]
\centering
\caption{Conditional baselines for Task~2 (constrained generation). Top: retrieval methods; bottom: learned generative models with bounding-box outputs. FID is \emph{partition-render} FID (Section~\ref{sec:task2_protocol}). $^\dagger$Room counts are fixed by conditioning.}
\label{tab:gen_baselines}
\small
\begin{tabular}{@{}lcccc@{}}
\toprule
\textbf{Method} & \textbf{Room Count Acc.} & \textbf{Adj. Satisfaction} & \textbf{Boundary IoU} & \textbf{FID}$_\text{part.}$ \\
\midrule
Random retrieval         & 0.517 & 0.464 & 0.744 & -- \\
Count retrieval          & \textbf{0.996} & 0.684 & \textbf{0.984} & -- \\
Count + Graph retrieval  & 0.976 & 0.903 & 0.786 & -- \\
\midrule
CVAE                     & 0.995 & 0.683 & 0.795 & 25.9 \\
CVAE + Graph             & 0.995 & 0.679 & 0.801 & 25.0 \\
Layout GNN$^\dagger$      & 1.000 & \textbf{0.999} & 0.520 & 63.8 \\
\bottomrule
\end{tabular}
\end{table}

\begin{figure}[!htbp]
    \centering
    \includegraphics[width=\linewidth]{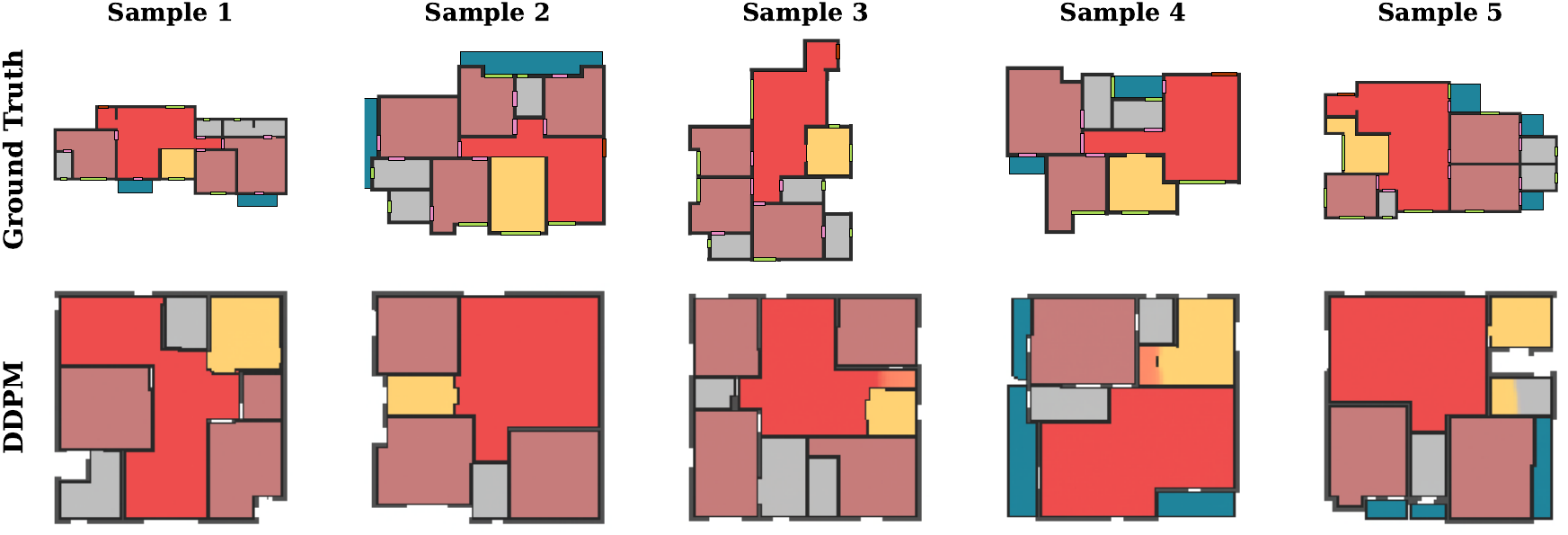}
    \caption{Sampled DDPM outputs for five held-out targets. Outputs are visually realistic (FID~6.1), with openings implicit.}
    \label{fig:qualgen}
\end{figure}

\paragraph{Results.}
Count retrieval achieves 99.6\% room-count accuracy; graph reranking improves adjacency satisfaction to 90.3\%. Under partition-FID, the layout GNN reaches near-perfect adjacency (0.999) but high FID (63.8) because it forces rectangular rooms. Both CVAEs keep 99.5\% room counts but only $\sim$0.68 adjacency, collapsing the living room into a dominant rectangle (Figure~\ref{fig:qualgen}) and yielding FIDs of 25.9/25.0. Unconditionally, the DDPM reaches FID$_\text{raster}$~6.1 against natively rasterized plans. Pixel generation thus gives visual fidelity without structural guarantees, while conditioned bbox methods enforce counts at the cost of spatial coherence.

\paragraph{Ablation: do typed edges help?}\label{sec:ablation_typed}
ResPlan exposes four edge types (\textit{via\_door}, \textit{adjacency}, \textit{direct}, \textit{via\_window}). CVAE+Graph collapses them into a 15-d type-pair count vector; we compare this with a 60-d \emph{typed} encoding (15 type-pair counts $\times$ 4 edge types) under the same best-of-20 sampling protocol. Table~\ref{tab:ablation_typed} shows the typed variant matches the untyped baseline (FID 25.0 vs.\ 25.0; adjacency 0.590 vs.\ 0.592; IoU 0.553 vs.\ 0.555). At this model scale, the limiting factor is the CVAE's dominant-living-room mode rather than graph granularity, but the released typed edges enable relational decoders and per-edge-type message passing~\citep{schlichtkrull2018rgcn}.

\begin{table}[t]
\centering
\caption{Typed vs.\ untyped graph conditioning for CVAE+Graph (Task~2). Same architecture, training schedule, augmentation, and best-of-20 sampling protocol. FID is partition-render FID against partition-rendered real plans. Adjacency and IoU are recomputed under the shared protocol and therefore differ slightly from Table~\ref{tab:gen_baselines}.}
\label{tab:ablation_typed}
\small
\begin{tabular}{@{}lcccc@{}}
\toprule
\textbf{Graph conditioning} & \textbf{Cond.\ dims} & \textbf{FID}$_\text{part.}$ & \textbf{Adj.\ Sat.} & \textbf{Boundary IoU} \\
\midrule
Untyped (15-d aggregate)              & 25 & 25.0 & 0.592 & 0.555 \\
Typed (60-d, 4 edge-types $\times$ 15) & 70 & 25.0 & 0.590 & 0.553 \\
\bottomrule
\end{tabular}
\end{table}

\begin{figure}[t]
    \centering
    \includegraphics[width=\linewidth]{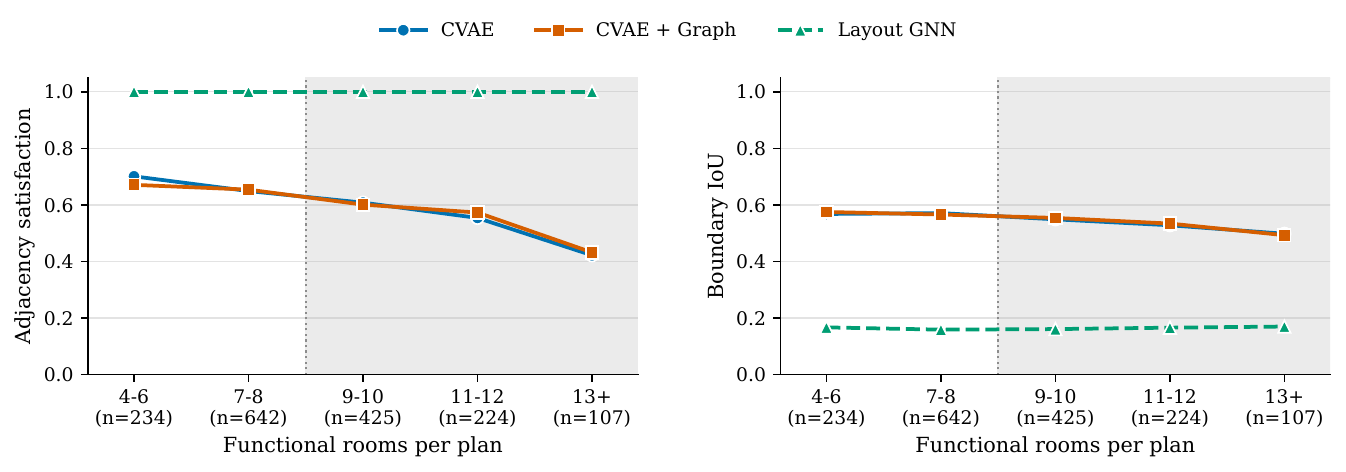}
    \caption{Adjacency satisfaction and boundary IoU vs.\ functional-room-count bucket. Shaded 9+ buckets are outside converted RPLAN support; CVAEs degrade with size, while Layout GNN keeps adjacency but low IoU.}
    \label{fig:scaling}
\end{figure}

\paragraph{Scaling with plan complexity.}\label{sec:scaling}
To probe scaling with layout complexity, the 1{,}632 test plans are bucketed by functional-room count; Figure~\ref{fig:scaling} reports per-bucket adjacency satisfaction and boundary IoU. Both CVAEs degrade monotonically: adjacency drops from 0.70 in the 4--6-room bucket to 0.42 in the 13+-room bucket, and IoU from 0.57 to 0.50. The drop starts within the converted RPLAN split's observed support (7--8 rooms) and accelerates in the 9+-room regime absent from that split (shaded). The Layout GNN satisfies adjacency by construction (1.000 across buckets) but remains at low IoU ($\sim$0.16), failing to fit the boundary as layouts grow.

\begin{figure}[!htbp]
    \centering
    \includegraphics[width=\linewidth]{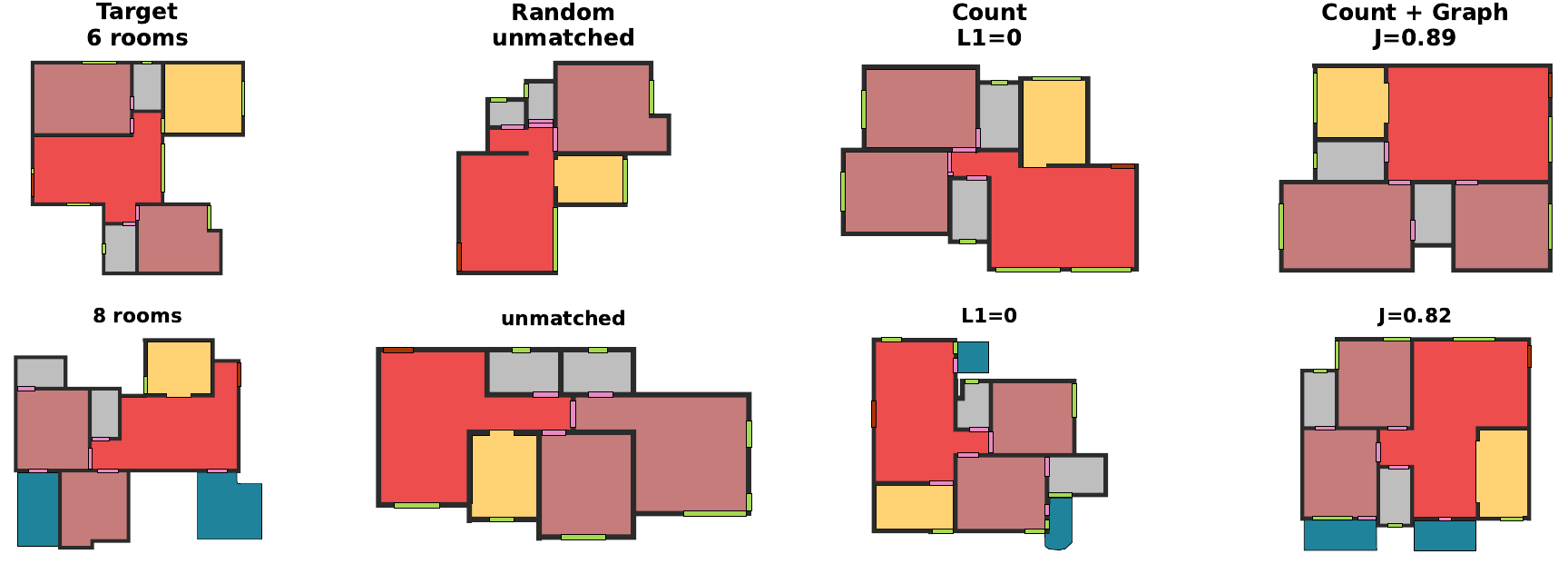}
    \caption{Qualitative retrieval for Task~2. Count+Graph retrieval preserves functional adjacencies better than count-only matching, showing why typed connectivity is useful beyond room counts.}
    \label{fig:retrieval}
\end{figure}

\subsection{Task~3: Plan-to-Graph Extraction}
Given only the polygonal layout, the goal is to recover the typed connectivity graph. We evaluate four heuristic edge detectors (proximity, shared boundary, complete graph, and an oracle upper bound) followed by an edge-type classifier that distinguishes \textit{via\_door}, \textit{adjacency}, \textit{direct}, and \textit{via\_window}. Two classifiers share the same 47 geometric features (shared-boundary length, opening overlap, polygon-pair distance, area ratios, axis-aligned offsets): a gradient-boosted (GB) tree ensemble and an inductive GraphSAGE node-conditioned variant~\citep{hamilton2017graphsage}. Shared-boundary detection alone recovers 98.2\% of true edges (F1 0.971), confirming that adjacency in ResPlan is overwhelmingly mediated by polygon contact rather than long-range visibility. Adding the GB classifier on top raises type accuracy from the 50.4\% majority-class baseline to 85.0\% (Table~\ref{tab:plan2graph}); the SAGE variant reaches 83.1\% with macro-F1 0.782, indicating room for richer relational architectures. Remaining errors concentrate on the rare \textit{via\_window} class (3.0\% of edges), which is most often confused with \textit{adjacency} when the window straddles a shared wall.

\begin{table}[!tbp]
\centering
\caption{Plan-to-graph extraction baselines (Task~3). Type accuracy is on correctly detected edges; \textsuperscript{$\dagger$}GB, \textsuperscript{$\ddagger$}SAGE.}
\label{tab:plan2graph}
\small
\begin{tabular}{@{}lcccc@{}}
\toprule
\textbf{Method} & \textbf{Precision} & \textbf{Recall} & \textbf{F1} & \textbf{Type Acc.} \\
\midrule
Proximity            & 0.564 & 0.906 & 0.695 & 0.504 \\
Shared boundary      & 0.961 & 0.982 & 0.971 & 0.504 \\
Shared boundary + GB\textsuperscript{$\dagger$} & 0.961 & 0.982 & 0.971 & \textbf{0.850} \\
Shared boundary + SAGE\textsuperscript{$\ddagger$} & 0.961 & 0.982 & 0.971 & 0.831$^{*}$ \\
Complete graph       & 0.315 & 1.000 & 0.480 & 0.504 \\
Oracle (GT)          & \textbf{1.000} & \textbf{1.000} & \textbf{1.000} & 1.000 \\
\bottomrule
\end{tabular}\\
\footnotesize
$^{*}$Macro-F1 0.782 for the learned SAGE edge classifier.
\end{table}

\FloatBarrier

%------------------------------------------------------------------------
\section{Discussion and Conclusion}
\label{sec:discussion}
%------------------------------------------------------------------------
ResPlan provides 17{,}000 unit-level vector floor plans paired with typed connectivity graphs and metric coordinates, with reproducible baselines for constrained generation, room labeling, and plan-to-graph extraction. Three properties distinguish it: (i)~native unit-level geometry without dwelling-extraction from larger floor plates; (ii)~four typed relations (\textit{via\_door}, \textit{via\_window}, \textit{adjacency}, \textit{direct}) separating door- from window-mediated access; and (iii)~broad layout complexity, with 34.6\% of held-out plans extending past the converted RPLAN eight-room support (Figure~\ref{fig:coverage}).

Task~2 reveals a gap between bounding-box generators that respect counts but collapse the living room and graph-conditioned methods that satisfy adjacency without realistic geometry; the scaling curves (Figure~\ref{fig:scaling}) localize this to the 9+-room regime outside RPLAN support. Cross-regional transfer (Section~\ref{sec:cross}) and the typed-edge ablation (Section~\ref{sec:ablation_typed}) suggest the next gains will come from joint geometry--graph models and per-edge-type message passing.

\subsection{Limitations}
\label{sec:limitations}
\paragraph{Geographic and cultural scope.} Every plan originates from South Asian residential listings. Room-type distributions, unit sizes, and layout conventions therefore reflect one regional tradition, and models trained on ResPlan should not be assumed to transfer to Western or East Asian residential architecture. Section~\ref{sec:cross} quantifies rather than excuses this, with substantial accuracy drops in both directions against RPLAN and MSD. Extending coverage to other markets is the primary direction for a future release.

\paragraph{Representation scope.} Plans are single-floor and unit-level: multi-storey circulation, furniture, and 3D information are absent. Wall thickness is normalized per plan, so within-plan variation between structural walls and thin partitions is not preserved (99.3\% of plans fall in the 10--40\,cm range). Applications requiring true structural wall geometry should treat this as an approximation.

\paragraph{Annotation artifacts.} Room boundaries are recovered by contour tracing, and a small tail of polygons retains jagged, stair-stepped contours rather than clean architectural edges: 0.52\% of room polygons exceed 30 vertices and 0.02\% exceed 100. Automated structural checks over the full dataset bound the remaining quality: doors align to a wall band in 99.94\% of cases and connect at least two rooms in 99.34\%, windows align in 99.98\%, and the summed room area falls within $\pm$25\% of the listed gross area for 98.7\% of plans. These are geometric checks and do not establish semantic room-label accuracy, for which the evidence remains the stratified 500-plan manual audit.

\paragraph{Near-duplicate plans.} Because listings are sometimes republished, the dataset contains repeated layouts. A geometry-based scan (label-raster IoU maximized over the eight dihedral symmetries, threshold 0.90) identifies 1{,}170 redundant plans (6.9\%) in 931 clusters, and 154 of the 1{,}632 test plans have a near-duplicate in the training split. The effect on benchmark results is small---removing every affected test plan changes Task~1 accuracy by $-0.15$~pp---but a deduplication-aware split is distributed alongside the canonical split so that users can avoid the leakage entirely.

\paragraph{Provenance.} Source platform identities are withheld to comply with platform access terms, and the collection and OCR/vectorization scripts are not redistributed. This limits independent auditing of geographic sampling and of the conversion pipeline itself, which we accept as the cost of honoring those terms.

\bibliographystyle{plainnat}
\bibliography{citations}

\end{document}